\documentclass[conference]{IEEEtran}
\IEEEoverridecommandlockouts
\usepackage{cite}
\usepackage{amsmath,amssymb,amsfonts}
\usepackage{algorithmic}
\usepackage{graphicx}
\usepackage{hyperref}
\usepackage{textcomp}
\usepackage{xcolor}
\def\BibTeX{{\rm B\kern-.05em{\sc i\kern-.025em b}\kern-.08em
    T\kern-.1667em\lower.7ex\hbox{E}\kern-.125emX}}
\begin{document}

\title{Efficient Yet Deep Convolutional Neural Networks for Semantic Segmentation\\
}

\author{\IEEEauthorblockN{Sharif Amit Kamran}
\IEEEauthorblockA{\textit{Center for Cognitive Skill Enhancement} \\
\textit{Independent University Bangladesh}\\
Dhaka, Bangladesh \\
sharifamit@iub.edu.bd}
\and
\IEEEauthorblockN{Ali Shihab Sabbir}
\IEEEauthorblockA{\textit{Center for Cognitive Skill Enhancement} \\
\textit{Independent University Bangladesh}\\
Dhaka, Bangladesh \\
asabbir@iub.edu.bd}
}

\maketitle

\begin{abstract}
Semantic Segmentation using deep convolutional neural network pose more complex challenge for any GPU intensive task. As it has to compute million of parameters, it results to huge memory consumption. Moreover, extracting finer features and conducting supervised training tends to increase the complexity. With the introduction of Fully Convolutional Neural Network, which uses finer strides and utilizes deconvolutional layers for upsampling, it has been a go to for any image segmentation task. In this paper, we propose two segmentation architecture which not only needs one-third the parameters to compute but also gives better accuracy than the similar architectures. The model weights were transferred from the popular neural net like VGG19 and VGG16 which were trained on Imagenet classification data-set. Then we transform all the fully connected layers to convolutional layers and use dilated convolution for decreasing the parameters. Lastly, we add finer strides and attach four skip architectures which are element-wise summed with the deconvolutional layers in steps. We train and test on different sparse and fine data-sets like Pascal VOC2012, Pascal-Context and NYUDv2 and show how better our model performs in this tasks. On the other hand our model has a faster inference time and consumes less memory for training and testing on NVIDIA Pascal GPUs, making it more efficient and less memory consuming architecture for pixel-wise segmentation.
\end{abstract}

\begin{IEEEkeywords}
Deep Learning, Convolutional Neural Network, Semantic Image Segmentation, Skip Architectures
\end{IEEEkeywords}

\section{Introduction}
With the introduction of convolutional neural network, the image recognition task has accelerated with great pace and has given state-of-the-art results for classification, detection and semantic segmentation alike. Classification is to recognize the the whole image to a certain class. Whereas in detection, each object has to be identified with a bounding box accurately. For semantic segmentation every pixel of the object in the image has to be classified to a corresponding class. Over the years many classification models yielded better results \cite{vgg,imagenet,goingdeep} in their independent task. Due to this success they have also been used as a base model for acing in extracting local features and giving finer output \cite{fcn,dilation} for semantic segmentation tasks.

Problem definition of the task in hand is to keep the global structure in contrast with the local context \cite{fcn,dilation}. Here, the global structure means the shape of the objects as a whole and how they are placed in the image with respect to other objects. On the other hand, the local features means the small geometric shapes like the sharp edge, circles etc. For example, if we consider the bipedal humans as the object and its shape as the global structure, then the shape of the eyes, nose, color of the lips can be considered as the local features. Most of the time the local features tend to get lost while training the neural networks and global context seems to dominate throughout the segmentation mask. So for extracting those local fine features, Skip architectures were introduced \cite{fcn,dilation,rgb-d,crfasrnn} with the preexisting segmentation architecture. This fine output from skip connections were then element-wise summed with the coarse semantic information on the top most layers of the neural net. By using Skip architectures the image representation becomes finer and less coarse.

The drawback for designing convolutional neural network with high level computing for pixel-wise classification seems to be the huge amount of memory required for the task. Firstly, orthodox ConvNets have rather large receptive fields because of their convolutional filters and generates coarse blob-like output map when it is redefined to produce pixel-wise segmentation \cite{fcn}. Secondly, sub-sampling with max-pool in ConvNets diminishes the chance to get finer output\cite{fccrf}. Furthermore, similar labeling in neighboring pixels tends to get lost in deeper layers where the  upsampling\cite{deconvolution} takes place. So visual consistency and retaining spatial feature is one of the essential job for producing sharp segmentation mask. Falling short of producing such fine output can result in poor object portrayal and patch-like false regions in the segmentation mask \cite{fccrf,auto,imageparsing,asshier}.

Using finer strides \cite{fcn} and replacing vanilla convolution with dilated convolution \cite{dilation,parsenet} have shown better segmentation results while keeping the memory usage in check. Because with dilation the receptive fields can be increased exponentially\cite{dilation}. Whereas the filter of the convolution remains the same size as the previous filter. So with the expense of reducing the size of the filter and adding dilation between it, we can free up more memory for computing from the sixth convolutional layer in the architecture which is the most expensive layer.

Adding more Skip architectures seems to increase memory usage for the whole end-to-end network. But because additional memory has been freed up by using dilation\cite{dilation,parsenet}, extra skip connections can be added to upsample local features from other convolutional layers. In a feed forward network like Fully-Convolutional Neural Network\cite{fcn} (which is denoted by FCN) the size of the representation changes with each convolutions. As the structure is similar to an encoder-decoder network,the feature hierarchies from earlier layers have to be element-wise added with the upsampled \cite{fcn,deconvolution} layers in steps. 

Our proposal in this paper is an efficient yet deep feed forward neural net for a strongly supervised image segmentation task. Our work tends to integrate both dilated and vanilla convolution to recreate a FCN (which stands for Fully-Convolutional Neural Network) architecture which generates better output while consuming less memory. In addition, we introduce four skip architectures which fetches more local information lost in the network in bottom layers. These features are then upsampled and element-wise summed with the global feature map in steps ih the top layers. Which in turn produces better segmentation mask while keeping GPU memory consumption in check. Most Importantly, with this changes in architecture, the end-to-end deep network can be trained on any type of data while utilizing the usual back-propagation algorithm with more efficient and finer results.

\section{Literature Review}

Following section describe different procedures which has been proposed before for conducting semantic segmentation task using deep learning. Out of many approaches only few have been adopted for high computing pixel-wise segmentation. 

Our proposed model was developed based on a particular neural net that was used for image classification task \cite{imagenet,vgg,goingdeep} and the weights were transferred from it \cite{decaf,visual}. Transfer learning was seen being performed in classification task, afterwards it was applied to object detection tasks, lately it has been adopted for instance aware segmentation\cite{instance} and image segmentation models with a powerful classifier\cite{region,sbd,rgb-d}. We redesign and redefine the architecture and perform fine-tuning to get more sparse and accurate prediction for semantic segmentation. Furthermore. we compare different models with our one and show how it is more efficient and effective for semantic segmentation jobs.

Multi-digit recognition with neural network\cite{lenetextension}, an extension of LeNet\cite{lenet}, was such work where erratic range of values for input was first witnessed. Though the task was ordained for one dimensional data, Viterbi decoding was sufficient for such task. Three years later convolutional neural network was elongated for two dimensional feature output for processing postal address data \cite{postal}. These historical breakthroughs were designed to conduct small yet powerful detection task. Additionally LeCun et al.\cite{embryo} using fully convolutional inference developed a CNN for sparse multiple class segmentation of embryo. We have also seen FCNs being used in many recent deep layered nets for high level computation. Using Sliding window for integrated object detection and localization by Eigen et al. \cite{overfeat}, Recurrent neural network for scene labeling by Pinheiro et al. \cite{rnn} , and restoring dirt clad image using convNet by Eigen et al. \cite{imagerestoration} is such remarkable example. Training a FCN can be difficult, but has been used for detecting human parts and estimating pose efficiently by Tompson et al. \cite{humanpose}

Different approaches can be taken to get finer segmentation mask exploiting convolutional neural network. One such strategy could be to develop individual system for extracting dense features and detecting zoomed-in edges from images for finer semantic segmentation \cite{zoomout,regionparts}. A single step process can be, extract semantic feature with convnet and then using superpixels for figuring out the inner layout of the image. Another procedure can be to retrieve superpixels from the given image layout and then extracting features from images one by one\cite{zoomout,learnscene}. The only drawback of this approach is that the erroneous super pixels may result into fallacious predictions, irrespective how powerful feature extraction took place. Zheng et al.  \cite{crfasrnn,hornn} designed a RNN model and used Conditional random field to get finer features by training an end-to-end network for semantic segmentation. They also proposed a disjointed version of the same model having less accuracy and consuming more memory to prove that an end-to-end network always have an upper hand over two or even three stage effective segmentation retrieval procedure.

Another strategy could be to develop a model and train it using supervised image data and output the segmentation label map for each categories. Retaining the spatial information, one can replace the fully connected layers with convolutional layers in a deep convnet, which was shown by Eigen et al \cite{depthmap}. The most groundbreaking work so far was by Shelhamer and Long et al \cite{fcn} where the idea was, FCN can be designed to harness features to help classify pixel from the top-most layers, whereas the bottom layers can be used for detecting shapes,contour and edges. With element-wise summation of earlier layers with latter layers they introduced the idea of skip architecture. On the other hand conditional random fields was used to refine semantic segmentation furthermore\cite{crfasrnn,hornn}. CRF was also used by Snavely et al. \cite{material} and Chen et al. \cite{fccrf,weak} for refining the existing segmentation mask. Snavely et al. conducted recognition for materials and it segmentation, on the other hand Chen et al. developed better ways to obtain finer semantic image segmentation. Though the previous procedures included disjointed CRF for conducting post-processing on the segmented output, the method developed by Torr et al. \cite{crfasrnn,hornn} employed CRF as recurrent neural network and also developed higher order model which is an extension of CRFasRNN. Not only is the convnet is end-to-end but also it converges faster than the previous CRF models and produces finer segmentation mask.

Difference between dilated and vanilla convolution is the extra parameter called holes or dilation that affects the receptive fields of the convolution`s filter. The whole idea of Atrous algorithm, which is based on wavelet decomposition\cite{wavelet} is wholeheartedly based on dilated filter. In \cite{dilation} Fisher Yu et al. used the term ``dilated convolution`` instead of ``convolution with a dilated filter`` to formulate that no dilated filter weren't built or produced. Convolutional layer was modified instead to make way for a new parameter called dilation to alter the preexisting filter. In \cite{attention} Chen et al. made use of dilation to modify the architecture of Shelhamer et al \cite{fcn} to make it suitable for his task. In contrast, Yu et al. \cite{dilation} developed a new range of feed forward neural net which exploits dilated convolutions and multi-scale context aggregation but get rid of the preexisting skip architectures. 

\section{Segmentation Architecture}

\subsection{Transfer Learning from Classification Net}
VGGnet is a famous neural net which won ILSVRC14\cite{vgg} for image classification. The neural net worked on the principal of using $3 \times 3$ sized filters for feature extraction and concurrently joins each convolutions to make the receptive field bigger. We transferred weights from the VGG 19-layer network , removed the classifier from the network and turned all the fully connected layers to convolutions as done by Shelhamer et al.\cite{fcn}.

In covolutional neural network all the tensors have three dimensions of size $N\times H\times W$, where H and W are defined as height and width, and N is the color channel or feature output map. At first layer the image is taken as input, where the pixel size is $H \times W$, and the three color channel for RGB is N. As described by shelhamer et al. \cite{fcn} receptive fields is the locations in higher layers corresponds to the locations path-connected to the image.

\subsection{Spatial Information and Receptive Field}
The output feature map for each convolutions can be predefined, but the spatial dimension depends on the size of the filter, strides and padding. FCN architecture tends to keep the spatial dimension of each convolutions same before max-pooling (exception being Fc6 and Fc7 layers). Considering the output channel dimension be $O_{ij}$ and input channel dimension be $I_{ij}$ for any convolution layer, where $i,j$ is the spatial dimension. Equation \eqref{MPKeq1} can be used for getting the optimal output channel.
\begin{equation}
\label{MPKeq1}
O_{ij} = \frac{I_{ij} + 2P_{ij} - K_{ij}}{S_{ij}} + 1
\end{equation}
Here, $P$ stands for Padding, $K$ is for Kernel or filter size and $S$ is for stride. We choose a filter size of $3 \times 3$, single padding and stride 1 for convolutional layers. This helps us to retain the convolutional structure before pooling to reduce the spatial dimension.

For pooling we use stride of 2 and filter of $2 \times 2$, to lower the spatial length of the tensors. If we use the probable value for filter and stride, in equation \(1\), then we can see the output becomes half of the size of the input.

The tensors go through first convolution layer to second convolution layer and onward as it is fed forward through the net. In the first two convolution layers we have $3 \times 3$ filters. Therefore the first one has a receptive field of $3 \times 3$ and second one has receptive field of $5 \times 5$. From third to fifth set of convolutional layers we have four convolutions for each of them. So, the receptive field is quite larger than the earlier layers. Notice that receptive fields increases linearly after each convolutions. Let receptive field $r$, filter $f$ and stride be $s$. If the filter and stride values remains the same for concurrent convolutions, \eqref{MPKeg2} formulation can be used for computing the receptive field.
\begin{equation}
\label{MPKeg2}
r_{n} \times r_{n} = (2r_{n-1}+1) \times (2r_{n-1}+1) 
\end{equation}
where  $r_{n}$ is the receptive field of the next convolution and
 $r_{n-1}$ is the receptive field of the previous convolution.

The sixth layer which is the most expensive one comes next. It is a fully connected layer, having a spatial dimension of $4096 \times 4096$ with a filter size $7 \times 7$. The next is also a fully connected layer having a spatial dimension of $4096 \times 1000$ and a filter size of $1 \times 1$. We convert both of this layers to convolutional layers with filter size of $3 \times 3$ and $1 \times 1$ as done by Shelhamer et al.\cite{fcn}. 

\subsection{Decreasing Parameters using Dilation}
Fihser et al. \cite{dilation} combined dilation with vanilla convolution throughout the network and emphasized multi-context aggregation. Whereas, we stick to the original structure of FCN \cite{fcn} but include dilation only in the sixth convolution, which is the most expensive layer and has the most amount of parameter computation. Similar work was done before in Parsenet, by Liu et al. \cite{parsenet}, but they trained on a reduced version of VGG-16 net. We train on the original VGG-19 classification neural net and only change the filter size and enter dilation as parameter in Fc6 convolution (see Table \ref{table1}) while retaining the same number of parameters across all the convolutions. The relation between filter size and dilation can formulated using \eqref{MPKeg3}.
\begin{equation}
\label{MPKeg3}
K' = K+(K-1)(d -1)
\end{equation}
where K' is the new filter size, K is the given filter size and d is the amount of dilation.

A convolution with dilation 1 is the same as having no dilation.
Using equation \(3\), the filter size of the sixth convolution layer can be changed from $7 \times 7$ to $3 \times 3$ with a dilation size of 3 and filter size of 3. Fisher et al \cite{dilation} has also defined \eqref{MPKeq4} for calculating the receptive field of a dilated convolution considering the same condition as before.

\begin{equation}
\label{MPKeq4}
r_{i+1} \times r_{i+1} = (2^{i+2}-1) \times (2^{i+2}-1)
\end{equation}

\subsection{Deconvolution with Finer Strides}
\begin{figure}[tbp]
    \centering
    \includegraphics[width=8cm,height=8cm]{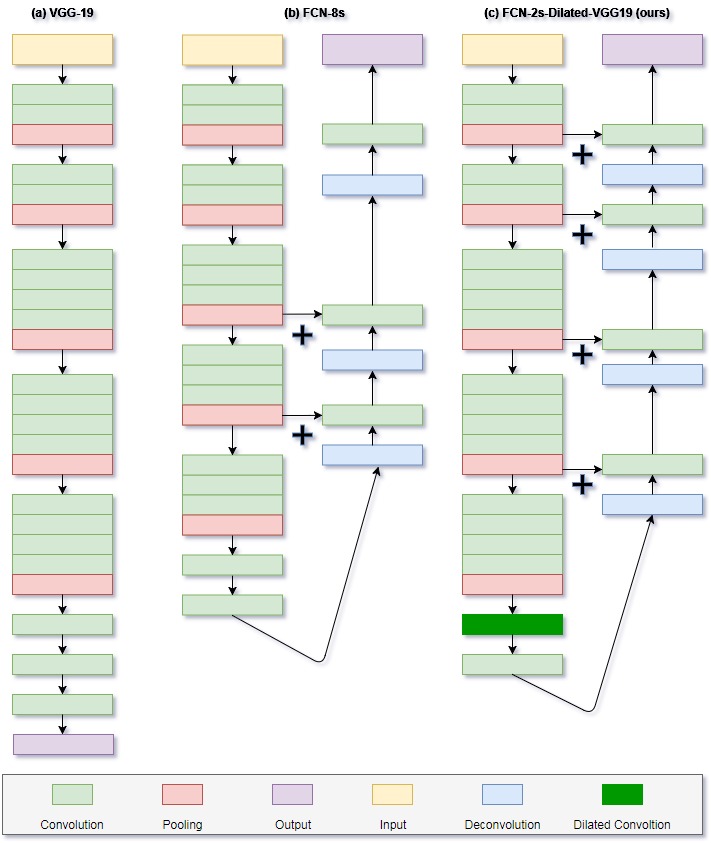}
    \caption{The comparision between VGG-16, FCN-8s and Dilated Fully Convolutional Neural Network with Skip Architectures. Dilated FCn-2s upsample stride 2 predictions back to pixels in five steps. The pool-4 to pool-1 are element-wise summed with stride 2, 4, 8 and 16 in steps in reverse order. This provides with finer segmentation and accurate pixel classification }
    \label{fig:skeleton}
\end{figure}

As upsampling with factor f means using convolution with
a fractional input stride of 1/f \cite{fcn}. If f is an integer,
we can reverse the forward and backward pass to make the upsampling work by replacing vanilla convolution with transpose convolution. So we use upsampling with finer stride as done by Shelhamer et al. \cite{fcn} for end-to-end calculation of pixel-wise semantic loss. But the author used strides of 32, 16 and 8, whereas we use a stride of 2 to upsample it in steps. This was done to element-wise summed  local features from the bottom layers using skip architectures (discussed in the next section).  Fig. \ref{fig:skeleton} shows the procedure in details.
In \cite{visual}, transpose convolution layers are called ``deconvolution layers``. The deconvolutional layers are used for ``bilinear interpolation`` as described in \cite{deconvolution} rather than learning. It was witnessed by Shelhamer et al.\cite{fcn} and Torr et al.\cite{crfasrnn} that 
upsampling in an end-to-end network is way faster and effective for learning dense prediction.

\subsection{Multiple Skip Architectures}
We adopt a similar procedure as fcn-8s-all-at-once \cite{fcn} for training rather than fcn-8s staged version  Because it is more time-consuming to do in stages while predicting with nearly the same accuracy as the all-at-once version. Moreover, for all-at-once version each skip architectures scales with fixed constant and these constants are chosen in such a way that it equals to the average feature norms across all skip architectures\cite{fcn}. This helps to decrease inference time a lot. For our case the inference time was less than 200ms for each forward and backward pass combined, whereas fcn-8s had an inference time of 500ms. We use a total of four skip architectures.  Also the skip architectures tends to consume less memory compared to the convolution layer due to its only operation being element-wise summation with other layer. 

\begin{table}[ht]
\centering
\caption{Parameters Comparison}
\begin{tabular}{|l|l|l|}
\hline
                      & FCN-8s       & Dilated FCN-2s (our)\\ \hline
inference time        & 0.5s         & 0.2s                \\ 
Fc6 Weights   & 4096x512x7x7 & 4096x512x3x3        \\ 
Dilation              & 1(none)      & 3                   \\ 
Fc6 Parameters        & 102,760,448  & 18,874,368          \\ 
Total Parameters      & 134,477,280  & 55,812,880             \\ \hline
\end{tabular}
\label{table1}
\end{table}

\section{Experiments}

\subsection{Metrics and Evaluation}
We use four different metrics to score the pixel-accuracy \eqref{pa}, mean-intersection-over-union denoted by mIOU \eqref{miou}, mean accuracy \eqref{ma} and frequency weighted accuracy denoted by fw-IU \eqref{fw}. As background pixels numbers in majority, pixel accuracy is not preferable.For semantic segmentation and scene labeling mean-intersection-over union is the most optimum choice for bench-marking.

\begin{equation}
\label{pa}
\textbf{pixel accuracy: }\frac{\sum_{i} N_{ii}}{\sum_{i}\sum_{j} N_{ij}}
\end{equation}
\begin{equation}
\label{ma}
\textbf{mean accuracy: }\frac{(1/N_{class})\sum_{i} N_{ii}}{\sum_{j} N_{ij}}
\end{equation}
\begin{equation}
\label{miou}
\textbf{mIOU: }\frac{(1/N_{class})\sum_{i} N_{ii}}{(\sum_{j} N_{ij} + \sum_{j} N_{ji} - N_{ii})}
\end{equation}
\begin{equation}
\label{fw}
\textbf{fw-IU: }\frac{(\sum_{k} \sum_{j} N_{kj})^{-1}\sum_{i} \sum_{j} N_{ij}  N_{ii}}{(\sum_{j} N_{ij} + \sum_{j} N_{ji} - N_{ii})}
\end{equation}

\normalsize where $N_{ij}$ is the number of pixels of class i predicted to belong to class j,
$N_{class}$ is the number of classes and $\sum_{j} P_{ij}$ is the total number of pixels of class i. The data was used as it is provided by \cite{pascal} and \cite{sbd}. No pre or post processing or augmentation was done to training or validation images to enhance the accuracy of the segmentation output. 

In Table \ref{table2}, Pixel Accuracy, Mean Accuracy, MeanIOU and FW Accuracy comparison between our model and Other FCN architecture. Both of our model outperforms the preexisting FCN structure for the reduced validation set.

\begin{figure}[tp]
    \centering
    \includegraphics[width=8cm,height=6cm]{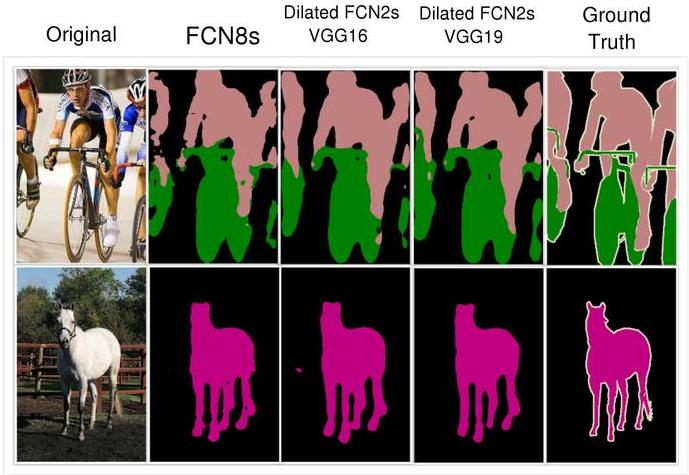}
    \caption{The Third and fourth image shows the output of our model. Fourth one being more accurate Dilated FCN-2s VGG19. The second image shows the output of the previous best method by Shelhamer et al.\cite{fcn}.}
    \label{fig:segmentationmask}
\end{figure}

\begin{table}[htbp]
\centering
\caption{Evaluation on PASCAL VOC2012 reduced validation set}
\label{table2}
\begin{tabular}{|l|l|l|l|l|}
\hline
Neural Nets                                                                  & \begin{tabular}[c]{@{}l@{}}Pixel \\ Accuracy\end{tabular} & \begin{tabular}[c]{@{}l@{}}Mean \\ Accuracy\end{tabular} & \begin{tabular}[c]{@{}l@{}}Mean \\ IOU\end{tabular} & \begin{tabular}[c]{@{}l@{}}FW \\ Accuracy\end{tabular} \\ \hline
FCN-8s-all-at-once                                                           & 90.8                                                      & 77.4                                                     & 63.8                                                & 84                                                     \\ \hline
FCN-8s                                                                       & 90.9                                                      & 76.6                                                     & 63.9                                                & 84                                                     \\ \hline
\begin{tabular}[c]{@{}l@{}}Dilated FCN-2s \\ using VGG16 (ours)\end{tabular} & 91                                                        & 78.3                                                     & 64.15                                               & 84.4                                                   \\ \hline
\begin{tabular}[c]{@{}l@{}}Dilated FCN-2s \\ using VGG19 (ours)\end{tabular} & 91.2                                                      & 77.6                                                     & 64.86                                               & 84.7                                                   \\ \hline
\end{tabular}
\end{table}

\subsection{Data-set and Procedure}
\subsubsection{\textbf{Pascal VOC}} Transfer learning was performed by copying weights separately  from VGG-19 and VGG-16 classification nets for our two models, Dilated FCN-2s-VGG19 and Dilated FCN-2s-VGG16. We adopt the Back propagation \cite{lenet} algorithm to train the network end-to-end with forward and backward pass.
We used dilation for our most expensive layer, Fc6 as seen in Table \ref{table1}. which reduced the number of parameters. Resulting into less computation by the machine yet faster inference time. The total time needed was 12 hours for both the networks to get the best mIOU using a single GPU. We used PASCAL VOC 2012 training data counting up to 1464 images. Validation was done on the reduced VOC2012 validation set of 346 images\cite{crfasrnn} in which we got 58 percent meanIOU.

\subsubsection{\textbf{Semantic Boundaries Dataset}} Extensive data was used to improve the pixel accuracy and mean-intersection-over-union score of both the models for which the training was done on Semantic Boundaries data-set \cite{sbd}. The set consists of 8498 training and 2857 validation data. Training was done on both the training and validation data summing up to 11355 images. The reduced set for validation was found by removing the common images in Augmented VOC2012 training set and VOC2012 validation set\cite{pascal}, resulting to 346 images. Table \ref{table2} to shows the comparative results of different FCN models. Our model, Dilated FCN-2s-VGG16 achieves a meanIOU of 64.1 percent and Dilated FCN-2s-VGG19 scores a meanIOU of 64.9 percent. Clearly, the deeper version of the model is more precise for pixel-wise-segmentation. Training was done with learning rate of $10e^{-11}$ with 200,000 iterations.

\begin{table}[bp]
\centering
\caption{PASCAL VOC 12 test results}
\label{table3}
\begin{tabular}{|l|l|}
\hline
Neural Nets                                                                  & MeanIOU  \\ \hline
FCN-8s   \cite{fcn}                                                                    & 62.2    \\ \hline
FCN-8s-heavy  \cite{fcn}                                                               & 67.2    \\ \hline
DeepLab-CRF \cite{fccrf}                                                                 & 66.4    \\ \hline
DeepLab-CRF-MSc  \cite{fccrf}                                                            & 67.1    \\ \hline
\begin{tabular}[c]{@{}l@{}}VGG19\_FCN \cite{vgg19fcn}\end{tabular}      & 68.1    \\ \hline
\begin{tabular}[c]{@{}l@{}}Dilated FCN-2s using VGG16 (ours)\end{tabular} & 67.6       \\ \hline
\begin{tabular}[c]{@{}l@{}}Dilated FCN-2s using VGG19 (ours)\end{tabular} & 69      \\ \hline
\end{tabular}
\end{table}

\subsubsection{\textbf{VOC2012 Test}} The test results as shown on Table \ref{table3} indicates our models scoring better than similar FCN architecture in Pascal VOC2012 Segmentation Challenge. We didn't train on any additional data, neither did we add any graphical model like CRF or MRF \cite{crfasrnn,hornn} to enhance the accuracy furthermore. Reason being it consumes $2\times$ more GPU memory for training. Moreover, our model scores better than FCN model in NYUDv2 sets too (see Table \ref{table5}). Fig. \ref{fig:segmentationmask} shows the segmentation mask compared to FCN-8s and the ground truth. Also Table \ref{table6} demonstrates, how less our nets consume memory for training and testing with GPU. As one can see, the reduction in memory usage is more than 20 percent for training with FCN-2s Dilated VGG16.

\begin{table}[tp]
\centering
\caption{Evaluation of Pascal Context data-set}
\label{table4}
\begin{tabular}{|l|l|l|l|l|}
\hline
Architectures                                                   & \begin{tabular}[c]{@{}l@{}}pixel\\ accu.\end{tabular} & \begin{tabular}[c]{@{}l@{}}mean\\ accu.\end{tabular} & \begin{tabular}[c]{@{}l@{}}mean\\ IU\end{tabular} & \begin{tabular}[c]{@{}l@{}}f.w.\\ IU\end{tabular} \\ \hline
O2P\cite{o2p2}                                                        & -                                                  & -                                                & 18.1                                             & -                                             \\ \hline
CFM\cite{o2p}                                                        & -                                                  & -                                                & 18.1                                             & -                                             \\ \hline
FCN-32s                                                         & 65.5                                                  & 49.1                                                 & 36.7                                              & 50.9                                              \\ \hline
FCN-16s                                                         & 66.9                                                  & 51.3                                                 & 38.4                                              & 52.3                                              \\ \hline
FCN-8s                                                          & 67.5                                                  & 52.3                                                 & 39.1                                              & 53.0                                              \\ \hline
CRFasRNN\cite{crfasrnn}                                                        & -                                                     & -                                                    & 39.28                                             & -                                                 \\ \hline
HO-CRF\cite{hornn}                                                          & -                                                     & -                                                    & 41.3                                              & -                                                 \\ \hline
\begin{tabular}[c]{@{}l@{}}DeepLab-\\ LargeFOV-CRF\cite{deeplab}\end{tabular} & -                                                     & -                                                    & 39.6                                              & -                                                 \\ \hline
\textbf{Ours}                                                   & 69.9                                                     & 54.9                                                    & 42.6                                              & 56.5                                                 \\ \hline
\end{tabular}
\end{table}

\subsubsection{\textbf{Pascal Context Data-set}} We train on more sparse data-set like Pascal Context which has 60 classes and pose more challenging pixel-wise prediction task \cite{context}. The data-set consists of 10103 images. We split the data set into 5105 validation images and rest are used as training set. Table \ref{table4} shows comparative results for Pascal Context Data-set. Our model, scores better mean-IOU of 42.6 percent than the other state-of-the-art models. Moreover, many deeper models with Higher Order CRF as post processing layer scored worse than our model. This clearly indicates that our model is better suited for pixel-wise prediction of sparse data-set. Training was done with learning rate of $10e^{-10}$ with 300,000 iterations.

\begin{table}[bp]
\centering
\caption{Evaluation on NYUDv2 data-set}
\label{table5}
\begin{tabular}{|l|l|l|l|l|}
\hline
                                                               & \begin{tabular}[c]{@{}l@{}}pixel\\ acc,\end{tabular} & \begin{tabular}[c]{@{}l@{}}mean\\ acc.\end{tabular} & \begin{tabular}[c]{@{}l@{}}mean\\ IU\end{tabular} & \begin{tabular}[c]{@{}l@{}}f.w.\\ IU\end{tabular} \\ \hline
Gupta et al. \cite{rgb-d}                                    & 60.3                                                 & -                                                   & 28.6                                              & 47                                                \\ \hline
FCN-32s RGB                                                    & 61.8                                                 & 44.7                                                & 31.6                                              & 46                                                \\ \hline

FCN-32s HHA                                                    & 58.3                                                 & 35.7                                                & 25.2                                              & 41.7                                              \\ \hline
\begin{tabular}[c]{@{}l@{}}FCN-2s Dilated RGB\end{tabular}   & 62.6                                                 & 47.1                                                & 32.3                                              & 47.5                                              \\ \hline

\begin{tabular}[c]{@{}l@{}}FCN-2s Dilated HHA\end{tabular}   & 58.8                                                 & 39.3                                                & 26.8                                            & 43.5                                              \\ \hline
\end{tabular}
\end{table}

\subsubsection{\textbf{NYUDv2 Data-set}} We train on NYUD version 2, an RGB-D dataset collected with the Microsoft Kinect. It consists of 1,449 RGB-D images, with pixel-wise semantic labels that is divided into 40 semantic classes by Gupta et al. \cite{nyud}. The data is split into 795 training images and 654 testing images. In, Table 5 the comparison of between fcn and our model is given. We train with Dilated FCN-2s VGG19 with three channel RGB images. Then we add depth information and train on a new model upgraded to take four-channel RGB-D input. Though the performance doesn't increase. Long et al. \cite{fcn} describes this phenomenon happens due to having similar number of parameters or the failure to propagate all the semantic gradients through the net. Following the footstep of Gupta et al. \cite{rgb-d}, we next train on three-dimensional HHA encoding of depth. The results proves to be more precise and yields better score for our model. Training was done with learning rate of $10e^{-10}$ with 150,000 iterations. Table \ref{table5} indicates comparative results for NYUDv2 data-set.

\begin{table}[bp]
\centering
\caption{GPU memory usage comparison}
\label{table6}
\scalebox{0.7}{
\begin{tabular}{cccccc}
\hline
\multicolumn{1}{|c|}{Models}                                                                          & \multicolumn{1}{c|}{\begin{tabular}[c]{@{}c@{}}GPU Memory\\ Usage\\ Training(MB)\end{tabular}} & \multicolumn{1}{c|}{\begin{tabular}[c]{@{}c@{}}GPU Memory\\ Usage\\ Training + Testing(MB)\end{tabular}} & \multicolumn{1}{c|}{\begin{tabular}[c]{@{}c@{}}Number\\ of\\ Parameters\\ (millions)\end{tabular}} & \multicolumn{1}{c|}{\begin{tabular}[c]{@{}c@{}}Inference\\ Time(ms)\end{tabular}} & \multicolumn{1}{c|}{\begin{tabular}[c]{@{}c@{}}Number\\ of\\ Classes\end{tabular}} \\ \hline
\multicolumn{1}{|c|}{Fcn-8s}                                                                          & \multicolumn{1}{c|}{3759}                                                                      & \multicolumn{1}{c|}{4649}                                                                                & \multicolumn{1}{c|}{134}                                                                           & \multicolumn{1}{c|}{500}                                                          & \multicolumn{1}{c|}{20}                                                            \\ \hline
\multicolumn{1}{|c|}{\textbf{\begin{tabular}[c]{@{}c@{}}Dilated Fcn-2s\\ VGG16(ours)\end{tabular}}}   & \multicolumn{1}{c|}{3093}                                                                      & \multicolumn{1}{c|}{4101}                                                                                & \multicolumn{1}{c|}{50.5}                                                                          & \multicolumn{1}{c|}{200}                                                          & \multicolumn{1}{c|}{20}                                                            \\ \hline
\multicolumn{1}{|c|}{\textbf{\begin{tabular}[c]{@{}c@{}}Dilated Fcn-2s\\ VGG19(ours)\end{tabular}}}   & \multicolumn{1}{c|}{3367}                                                                      & \multicolumn{1}{c|}{4309}                                                                                & \multicolumn{1}{c|}{55.8}                                                                          & \multicolumn{1}{c|}{200}                                                          & \multicolumn{1}{c|}{59}                                                            \\ \hline
\multicolumn{1}{|c|}{\begin{tabular}[c]{@{}c@{}}PascalContext\\ Fcn-8s\end{tabular}}                  & \multicolumn{1}{c|}{4173}                                                                      & \multicolumn{1}{c|}{5759}                                                                                & \multicolumn{1}{c|}{136}                                                                           & \multicolumn{1}{c|}{500}                                                          & \multicolumn{1}{c|}{59}                                                            \\ \hline
\multicolumn{1}{|c|}{\textbf{\begin{tabular}[c]{@{}c@{}}Dilated Fcn-2s\\ Context(ours)\end{tabular}}} & \multicolumn{1}{c|}{3975}                                                                      & \multicolumn{1}{c|}{5333}                                                                                & \multicolumn{1}{c|}{56.2}                                                                          & \multicolumn{1}{c|}{200}                                                          & \multicolumn{1}{c|}{59}                                                            \\ \hline
\multicolumn{1}{l}{}                                                                                  & \multicolumn{1}{l}{}                                                                           & \multicolumn{1}{l}{}                                                                                     & \multicolumn{1}{l}{}                                                                               & \multicolumn{1}{l}{}                                                              & \multicolumn{1}{l}{}                                                               \\
\multicolumn{1}{l}{}                                                                                  & \multicolumn{1}{l}{}                                                                           & \multicolumn{1}{l}{}                                                                                     & \multicolumn{1}{l}{}                                                                               & \multicolumn{1}{l}{}                                                              & \multicolumn{1}{l}{}                                                              
\end{tabular}
}
\end{table}

\begin{figure}[tp]
    \centering
    \includegraphics[width=8cm,height=8cm]{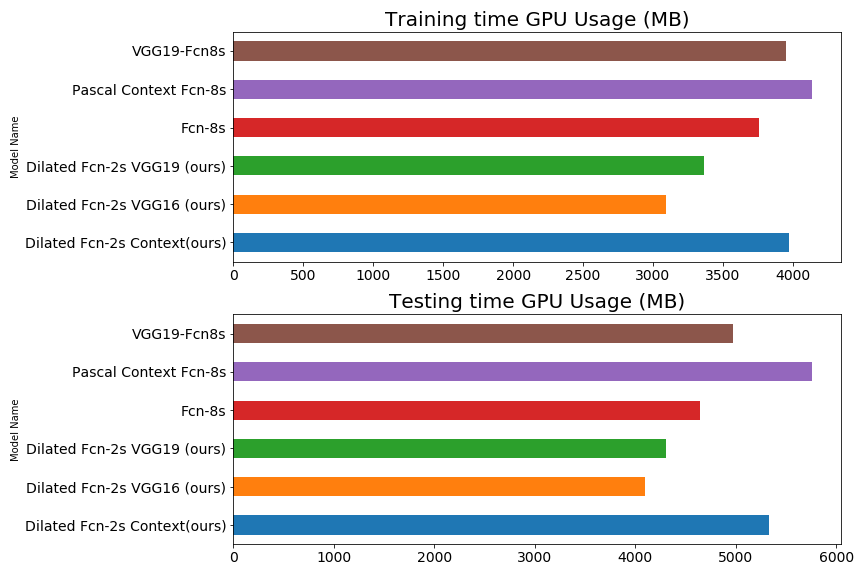}
    \caption{The difference in GPU usage while training and testing between our model and some FCN architectures }
    \label{fig:gpuusage}
\end{figure}

\subsection{Memory Efficiency}

In terms of memory efficiency, three of our architecture has shown remarkable results across different GPUs. As shown in Fig. \ref{fig:gpuusage}, training and testing time GPU usage has been decreased for all of the achitectures. Moreover, Dilated-Fcn2s-Vgg16 achieves better performance results by reducing 700 MB while training. For Dilated-Fcn2s-Context it has has shown slight improvement for both training and testing time while getting better even for more power consuming models.

In Table \ref{table6}, it can be seen that the GPU memory allocation for 20-class and 59-class segmentation task has been reduced by 10-20 percent. Additionally, the counterpart of Fcn-8s\cite{fcn} which is Dilated-Fcn2s, has shown remarkable results for both training and testing time GPU memory usage. On the other hand, the inference time required for three of our models are less than half of the other similar architectures. And for both 20 and 59 class segmentation it retains similar inference times.

\begin{figure}[tp]
    \centering
    \includegraphics[width=8cm,height=8cm]{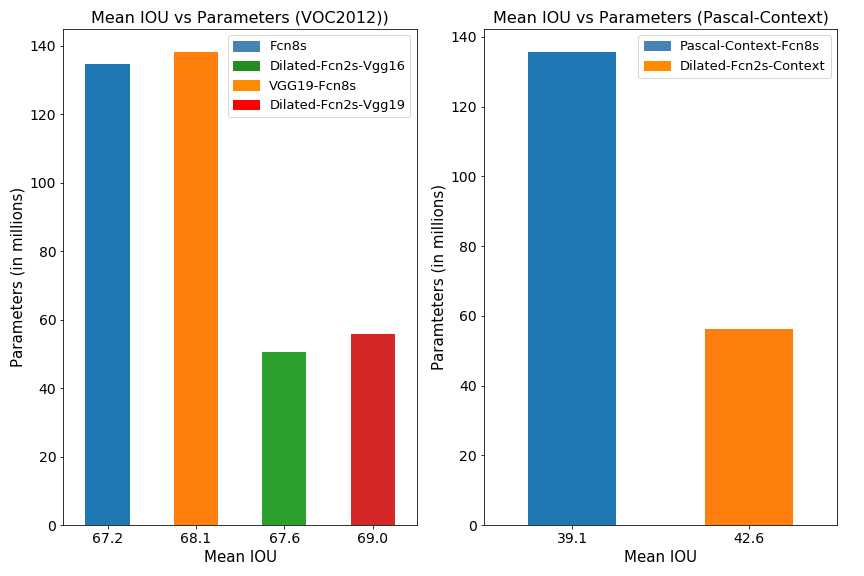}
    \caption{Number of parameters vs Mean IOU scores among different architectures }
    \label{fig:iouvsparam}
\end{figure}

Fig. \ref{fig:iouvsparam}, shows the comparisons of different models in regard to Mean intersection-Over-Union Vs. Parameters (in millions). The number of parameters of Fcn-8s \cite{fcn}, VGG19-Fcn \cite{vgg19fcn} architectures for training VOC2012 \cite{pascal} and Pascal-context \cite{context} data are in hundred of millions. Whereas, all three of our models require less parameters for calculation, hence less memory usage by both CPU and GPU. It brings to light another prospect that, huge number of parameters are not needed to increase accuracy or finer potrayal. Furthermore, wide receptive fields and training on sparse data can also effectively give better results.

All the models where trained and tested with Caffe \cite{caffe} on Nvidia GTX1060 and GTX1070 separately. The code for this model can be found at:  \url{https://github.com/SharifAmit/DilatedFCNSegmentation}

\begin{figure*}[tp]
    \centering
    \includegraphics[width=16cm,height=12cm]{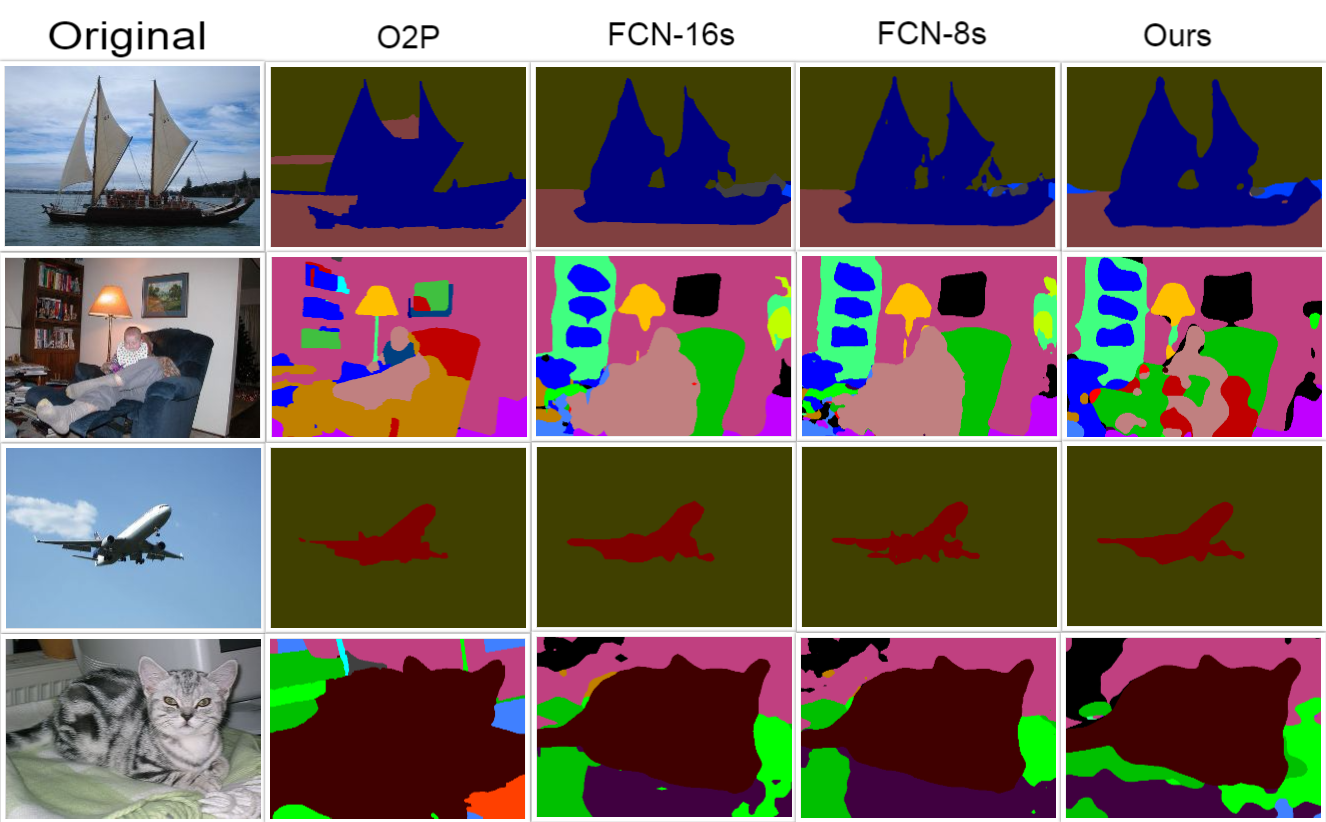}
    \caption{Results after testing on Pascal-Context dataset \cite{context} for 59-classes.The Fifth column of images show the output of our model which tends to be more accurate. On the second column O2P model's \cite{o2p} \cite{o2p2} output  which has wrongly predicted in many instances.}
    \label{fig:segmentationmask2}
\end{figure*}

\section{Conclusion}
Enhancing accuracy for pixel-wise segmentation requires huge amount of memory and time. Our benchmark result for PASCAL VOC2012 test data set set for 20 unique classes scored mean-IOU of 69 percent for Dilated-FCN-2s-VGG19 and 67.6 percent for Dilated-FCN-2s-VGG16. On the other hand, for sparse data-set like NYUDv2 for 39 unique classes and Pascal-Context for 59 unique classes our model scored pixel accuracy of 62.6 percent and mean IOU of 42.6 percent respectively. Fully convolutional networks can be used to transfer weights from pre-trained net, element-wise summing different layers to improve accuracy and to train end-to-end for entire images with extensive data. Dilation increases the receptive fields while decreasing parameters and inference time. The objective was to create efficient yet deep architectures for generating accurate output while using less computation resources. And the proposed models have remarkably produced accurate pixel-wise segmentation. Hopefully, this architectures can be further used for Semantic Segmentation tasks like Self-Driving cars, Medical Imaging and Robotics.

\section*{Acknowledgment}
We would like to thank Evan Shelhamer for providing the evaluation scripts and Caffe users community for their advice and suggestions. We also would like to acknowledge the technical support ``Center for Cognitive Skill Enhancement`` has provided to us.

\bibliographystyle{unsrt}
\bibliography{main}

\end{document}